# SemEval-2026 Task 4: Narrative Story Similarity and Narrative Representation Learning

**Hans Ole Hatzel[1], Ekaterina Artemova[2,3], Haimo Stiemer[4],**
**Evelyn Gius[4], Chris Biemann[1]**

University of Hamburg[1], Toloka AI[2], German UDS[3], TU Darmstadt[4]

{hans.ole.hatzel, chris.biemann}@uni-hamburg.de,
katya-art@toloka.ai, {haimo.stiemer,evelyn.gius}@tu-darmstadt.de

## Abstract

We present the shared task on narrative similarity and narrative representation learning – NSNRL (pronounced "nass-na-rel"). The task operationalizes narrative similarity as a binary classification problem: determining which of two stories is more similar to an anchor story. We introduce a novel definition of narrative similarity, compatible with both narrative theory and intuitive judgment. Based on the similarity judgments collected under this concept, we also evaluate narrative embedding representations. We collected at least two annotations each for more than 1,000 story summary triples, with each annotation being backed by at least two annotators in agreement. This paper describes the sampling and annotation process for the dataset; further, we give an overview of the submitted systems and the techniques they employ. We received a total of 71 final submissions from 46 teams across our two tracks. In our triple-based classification setup, LLM ensembles make up many of the top-scoring systems, while in the embedding setup, systems with pre- and post-processing on pretrained embedding models perform about on par with custom fine-tuned solutions. Our analysis identifies potential headroom for improvement of automated systems in both tracks. The task website includes visualizations of embeddings alongside instance-level classification results for all teams.[1]

## 1 Introduction

In this shared task, we seek to advance the computational modeling of narrative similarity. We understand narrative similarity to refer to the perception of story relatedness, focusing on abstract patterns of causality and progression rather than concrete details (e.g., names, actors, objects, settings). The submitted systems evaluate narrative similarity either in a direct-comparison setup or via distances

---

[1]https://narrative-similarity-task.github.io/results/

⚓ Anna loses her purse. She retraces her steps but cannot find it. Dan finds it and helpfully returns it to her.

**A**: Brian lost his backpack. He was terrified because there were important documents in it. After an hour of intense search he finally found it.

**B**: Alex lost his engagement ring while swimming. After hours of looking, he still can not find it. Karen finds the ring while magnet fishing and returns it.

Figure 1: Annotators are asked to choose which of two texts A or B is narratively more similar to the anchor text. Here we show an example from our annotation guidelines with the reference solution **B**.

between dense representations. Rather than using full stories, we focus on synopses extracted from Wikipedia articles. This approach has two main motivations: (1) shorter texts make the task more computationally approachable, and (2) under limited annotation resources, considering only summaries allows us to cover more stories. We argue that the aspects central to narrative similarity are typically captured in plot summaries, though some summaries may lack sufficient detail to cover all story similarity aspects.

Our task features two setups, which we refer to as tracks. In **Track A**, systems assess the narrative similarity in a direct-comparison setup, choosing which of two candidate stories is more similar to a reference (*anchor*) story. In **Track B**, on the other hand, systems create narrative representations in the form of embeddings. These representations are tested in the same manner as for Track A: embedding distances should adhere to the partial similarity orderings annotated for individual summaries.

This shared task evaluates automated systems against human baselines to validate both our novel definition of narrative similarity and the effective-ness of modern narrative representations. We structure the work as a shared task to encourage a wide range of approaches, be they grounded in narrative theory or motivated by technological advances.

## 2 Related Work

Existing narrative similarity datasets, such as the one by Chaturvedi et al. (2018) and Tell-Me-Again (Hatzel and Biemann, 2024b), are not based on human judgments but instead rely on external similarity information. Specifically, Chambers and Jurafsky (2009) rely on lists of movie remakes while Hatzel and Biemann (2024b) mine story summaries across language versions of the Wikipedia. Our own retelling dataset (Hatzel and Biemann, 2024a) is essentially compiled from existing lists of retellings using LLM prompting.

Fisseni and Löwe (2012) asked annotators to rate pairs of Dutch folk tales with regard to their similarity on a scale of 1 to 5. They approach candidate selection by sampling stories with identical story types and/or genres, in addition to employing lexical similarity measures. They also let annotators provide motivations for their decisions and grouped them into categories of narrative similarity, such as *Character*, *Genre*, *Motifs*, or *Structure*. The dataset is not publicly available. Chen et al. (2022) annotate similarity on news articles; like Fisseni and Löwe (2012), they use a 5-point scale, specifically annotating narrative similarity as one of the eight dimensions of similarity.

To our knowledge, all prior works in the story similarity domain utilize scalar similarity ratings. Scalar similarity has several advantages: annotators only need to evaluate one pair at a time, and a scalar notion can provide an ordering for multiple texts without relying on multiple pairwise comparisons. On the other hand, it is difficult to define points on a scale that annotators follow precisely. Numerically small scales also provide limited discriminative value, while larger ones are hard to annotate consistently (Miller, 1956). Contrastive annotation setups, long established as a method (Thurstone, 1927), can address these limitations by more accurately modeling similarity relationships and are conceptually simpler to reason about than scales (e.g., we can use accuracy rather than correlation). Kiritchenko and Mohammad (2017) have explored the general case of this consideration, comparing Likert-scale-style ratings with best-to-worst orderings of $n$ items. Other recent work has also found contrastive setups to work well, for example, in relevance judgments (Carterette et al., 2008). Our triple-based approach–essentially a Best-Worst Scaling variant–parallels the pairwise preference tuning used in LLM post-training setups (Jiang et al., 2025).

### 2.1 Symbolic Representations of Narrative

The symbolic representation of narratives has received considerable attention through concepts such as narrative chains and schemas (Chambers and Jurafsky, 2009; Finlayson, 2012) and affect state networks (Lehnert, 1981). Here, a narrative is represented in a structured form, such as by its constituent events. Assessing the narrative similarity has also long been pursued, using the aforementioned symbolic representations and other features (Miller et al., 2015; Chaturvedi et al., 2018). With advances in document encoder models, it has become feasible to use representation learning in the context of narratives, automatically inferring what was previously explicitly engineered as story features. Embeddings obtained in this way allow for corpus-scale analysis and retrieval based on narrative similarity (Hatzel and Biemann, 2024a; Sterner et al., 2026).

## 3 Towards A Definition of Similarity

None of the aforementioned structural approaches provides a clear model of narrative similarity, although edit-distance measures could theoretically be applied to their representations. In this work, we do not seek to operationalize a specific understanding of stories, as there is a wide range of conceptualizations that appear equally valid but emphasize different aspects of stories. Therefore, our guidelines are designed to balance a formal, consistently applicable definition of story similarity with one that aligns with human intuition, while placing no undue focus on any specific aspect of narrative.

To guide our annotators' decisions, we define three categories for judging similarity, all relevant to narrative analysis.

- **Course of Action**: The sequence of happenings in the story.

- **Outcomes**: The results of the happenings in the story, excluding intermediate results that change later on.

- **Abstract Theme**: The motifs and themes explored in the story. This aspect does not cover the concrete setting of a story.

We ask annotators to consider these three aspects when judging story similarity, but intentionally do not provide specific guidance on how to weigh them. Our guidelines, including the annotator training examples, are available online (Hatzel et al., 2025).

## 4 Task Structure

The task features two tracks; participants may submit to either or both.

**Track A: Comparative Narrative Similarity**
Participants are given a story triple: an anchor story $a$ and two candidate stories $c_1$ and $c_2$. Systems need to decide which of the two candidate stories, from a narrative perspective, is more similar to the anchor. Figure 1 shows our annotation setup, which is analogous to the classification setup in Track A.

**Track B: Narrative Representation Learning**
In this track, participants are given a collection of individual stories and produce an embedding representation for each. We evaluate embeddings using cosine distance: embeddings of narratively similar stories should be close in embedding space.

Evaluation in Track B uses the same triple annotations as Track A. If the more similar candidate is closer to the anchor in the embedding space than the corresponding distractor, the system is correct with respect to this triple. We verify that embedding distances for annotated triples align with human judgments, ignoring unannotated pairwise similarities. Formally, we compare $\text{dist}(a, c_1)$ and $\text{dist}(a, c_2)$, where dist denotes the cosine distance. We do not release the composition of the triples ahead of time; instead, we only provide individual stories.

## 5 Dataset

For the shared task, we collect 1,000 triples for the development and evaluation sets, each annotated by at least two annotators. The *dev*, *train*, and *sample* splits are drawn from the same distribution, so we expect performance differences only due to random variation. See Table 1 for an overview of the splits. In addition, in place of a training dataset, we provide a larger set of synthetically created samples (see Section 5.4 for details). Some stories occur in multiple triples; as a result, Track B contains fewer stories than three times the number of triples. Note that for all splits except the test split, Track A and B are based on the same annotations, meaning that the annotated triples in Track A correspond exactly to the individual stories in Track B. For the test split in Track B, we withhold the makeup of the evaluated triples to make it impossible to choose ideal embeddings based on individual triple judgments.[2]

| Split | Track A (Triples) | Track B (Stories) |
|---|---:|---:|
| Sample | 39 | 108 |
| Dev | 200 | 479 |
| Test | 400 | 849 |
| **Total** | 1,039 | 1,436 |
| Synthetic | 1,900 | 5,700 |

Table 1: For all but the test data, we use the same stories for Track A and B. For Track A, we list the number of items and, for Track B, the number of individual triples.

There are three stages to our dataset creation: (1) summary sourcing, (2) candidate triple sampling, and (3) human annotation.

### 5.1 Summary Sourcing

We source our stories from the English portion of Tell-Me-Again (Hatzel and Biemann, 2024b). We first filter the dataset for lengths from four to eight sentences. Including longer summaries, while certainly worthwhile, would require vast annotation resources, as Wikipedia summaries tend to be very information-dense; understanding longer summaries without knowledge of the source work takes considerable effort. Under resource constraints, we elect to annotate many short stories rather than fewer longer ones. After length filtering, we prompt gpt4o-mini[3] to remove summaries that contain only a premise or extraneous markup (e.g., headings). This prompt is detailed in Appendix B. We retain summaries that do not fully resolve the ending, as this is typical for short summaries and would otherwise dramatically reduce the number of suitable summaries.

In a step we refer to as rejection sampling, we prompt two different LLMs for each story triple, asking them whether A or B is more similar. We include only examples in which the two LLMs disagree, thereby allocating human annotation effort to difficult cases.

While we randomize the order of our candidate stories with respect to the sampling method, we do not randomize them across annotators. We observe a sizeable positional bias, with annotators leaning towards picking the first story as the more similar candidate. Annotators select the first story in approximately 58% of cases. For the purposes of the shared task, we randomize the order of the candidate stories once more. While this ultimately reduces data quality, it prevents systems from exploiting this bias.

---
[2]Of course, responses for all possible triple combinations could still be computed, but this would come at considerable computational cost.

[3]Specifically, we use gpt4o-mini-2024-07-18.

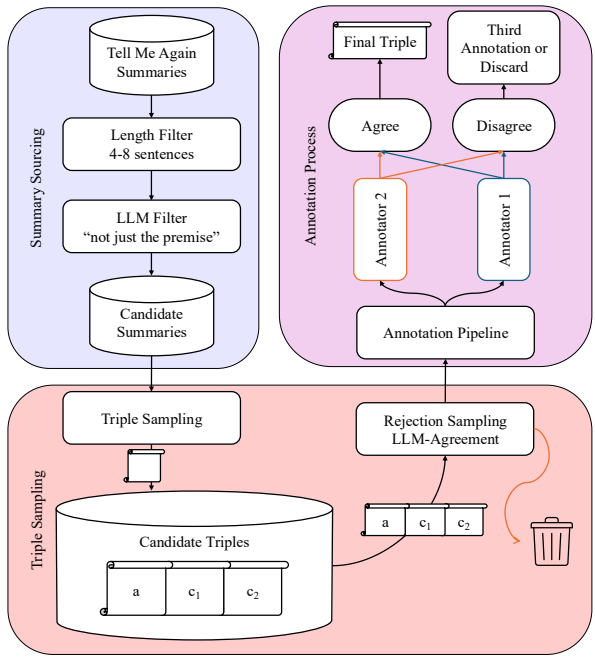

Figure 2: Our story sampling process begins with a set of English summaries. Only after multiple filtering and candidate sampling steps, do we present the resulting triples to our annotators.

|  | $\alpha$-Value | | |
| --- | --- | --- | --- |
|  | 2.0 | 3.0 | 5.0 |
| Agree | 5 | 8 | 10 |
| Disagree | 11 | 14 | 11 |
| Agree-Percentage | 68% | 63% | 52% |

Table 2: We showcase the impact of the $\alpha$ hyperparameter on a small sample of annotations. For the accuracy, we assume that both annotators agreeing indicates a correct prediction.

## 5.2 Candidate Triple Sampling

Intuitively, any pair of stories you pick from a dataset would, almost always, share very little similarity; as a result, annotating random samples yields few annotations of similar pairs. Chen et al. (2022) solve this problem by sampling related articles using Jaccard similarity and later using models trained on their initial annotation data, while Fisseni and Löwe (2012) address it using category metadata and lexical similarity.

Since we aim to identify cases of similarity irrespective of surface form, lexical similarity measures do not align with our objectives. Instead, we rely on our existing narrative model *story-emb* (Hatzel and Biemann, 2024a), selecting stories that, according to the model, are at least somewhat similar for annotation. As shown by Hatzel and Biemann (2024a) *story-emb* places less emphasis on named entities in its embedding representation as compared to its foundation model e5 (Ouyang et al., 2022). Yet, given that *story-emb* was trained on limited data, we expect it to retain some of the base model's characteristics. To control for this behaviour, we model similarity under *story-emb* as a linear combination of sentence similarity (as represented by similarity under the original e5 model) and story similarity. This is a simplifying assumption that empirically proves useful for our approach. Under this linear-combination assumption, we can remove the effect of e5 by subtracting its contribution.

We define the hyperparameters $\alpha \in R^+$ and $\beta \in [0, 1]$ to control the impact of this base-model correction.

$$sim_{\text{story}}(a, b) - sim_{\text{sentence}}(a, b)^\alpha \beta \quad (1)$$

As shown in Figure 2, we begin with a randomly sampled anchor story and select an assumed correct answer and an assumed distractor under our embedding model.[4] The random anchor story is sampled from the filtered stories, and candidates are drawn from the same pool.

The hyperparameters $\alpha$ and $\beta$ affect the similarity of candidates to a given anchor story. In addition, we employ the parameter $r$ to control the rank at which we sample the assumed distractor. This sampling process is visualized in Figure 3. For this task, we always take the most similar story as the assumed correct answer. We expect the difficulty to depend on the hyperparameters $\alpha$ and $\beta$ as well as the rank $r$. Large $\alpha$ values mean that only extreme cases of standard sentence similarity have an impact. In Table 2, we explore values for $\alpha$ on an early set of annotations. Although the sample size is small, we observe a consistent trend indicating that large $\alpha$-values adversely affect agreement. This data suggests that the annotators' agreement may be adversely impacted by (somewhat) similar entities in a story. We do not explore this parameter space; instead, we rely on $\alpha$-values that yield good agreement. While we performed some exploration in earlier annotation batches, we always kept $\beta$ fixed at 0.5 and roughly evenly sampled across $\alpha \in \{1.5, 2.0\}$ and $r \in \{0.98, 0.99\}$, meaning we sampled the distractor at the 98th or 99th most similar percentile. The agreement across two annotators in our data shares a minimal correlation with the rank (r=0.036) and the $\alpha$ value (r=-0.057).

---
[4]In practice, the embedding model often incorrectly picks a distractor that annotators consider to be more similar.

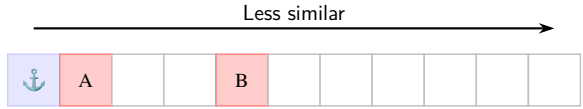

Figure 3: We sample the two options starting from the anchor and operating on a list of similar stories.

We note that the sampling process's reliance on *story-emb* means that comparing *story-emb* with other models on the dataset is potentially unfair.

After sampling triples, we employ an LLM filtering step, rejecting all those examples on which two commercial LLMs, in a single invocation of each model, already agree. Pilot annotations showed that without this step, our annotators were able to outperform LLMs by a small margin, but that LLMs were generally able to solve the task very well in a zero-shot setting.

### 5.3 Human Annotation & Annotation Quality

The annotation project was conducted on Toloka's platform with a total of 12 annotators, all of whom were native or near-native English speakers with extensive experience in annotation tasks; to mitigate potential bias, a daily cap was imposed on the number of tasks each annotator could complete.

We collected two or more annotations each for a total of 1,237 triples, all of which were filtered for LLM disagreement beforehand. In 831 cases, two annotators agree, making for roughly 83% of our final data. We observe annotator disagreement for 406 triples. For a subset of these (208 samples), we collected a third annotation to resolve the disagreement. This results in a total of 1,039 samples. We release 800 triples as the test set (400 per track) and use 200 triples as the development set.

On an individual annotation level, without taking into account the third annotation for a subset of the data, our agreement in terms of Krippendorff's Alpha is 0.33. This result is consistent with the roughly comparable annotations reported by Ehrmanntraut et al. (2022), who achieve Alpha values between 0.32 and 0.68 in a similar setup and across a range of dimensions. Although this agreement is relatively low, we want to emphasize that we are dealing with a highly subjective task and have already filtered the data to only include difficult cases.

Based on a pooled annotator and uniform task difficulty assumption, but taking into account the positional bias, we find each annotator to have an accuracy of 78% with respect to latent oracle labels (using essentially a simplified version of Dawid and Skene, 1979). Accordingly, for the data that is included based on exactly two annotations that agree, we expect an accuracy of almost 92% with regard to oracle labels. For the annotations where we adjudicate using a third annotator, we expect the accuracy of a single annotator (i.e., 78%). Overall, using the weighted average of the two cases, the theoretical accuracy of our released data with respect to oracle labels, under uniform difficulty and pooled-annotator assumptions, is 89%. In practice, these assumptions do not fully hold, so 89% accuracy should be regarded as an upper bound.

The individual aspects of similarity were not subject to the shared task, with the labels being held out until the end of the task. A priori, we expected relatively low agreement across the individual aspects and focused our monitoring through the annotation process on the similarity judgments, intending the similarity aspects primarily as a mental guide for annotators. The agreements (in terms of Krippendorff's Alpha) did end up being close to random chance for the abstract theme and the course of action (0.05 and 0.07) and slightly higher for the outcomes (0.11). We did not instruct annotators when to note each aspect, except by asking them to tick those aspects that they thought "significantly contribute to their decision". The poor agreement is likely attributable in part to the frequency with which the aspects are selected. While the outcomes were selected as having significantly contributed to the decision in 42.38% of cases, the abstract theme was selected in 98.60% of cases, with the course of action being selected in 69.53% of cases.

### 5.4 Synthetic Data

We create synthetic data intended as an alternative for a large-scale training dataset. Specifically, we prompt various commercial LLMs to first produce a story (in the style of a Wikipedia summary) and subsequently to produce one narratively similar and one narratively less similar story.

We employ two techniques for generating diverse stories, as early outputs proved repetitive. We create a number of seed topics to guide the story generation (e.g., *AI companionship*, *superhero origin*, or *love triangle*). The seed topic is inserted into the prompt using a placeholder. Additionally, we use a wide range of commercial models to produce both our stories and the seed topics. In this manner, we create 1,900 story triples.

### 5.5 Release Details

We release the test and development datasets with additional metadata, including the Wikidata ID of the work corresponding to each summary. All releases of our test set contain two canary GUID strings: one of our own to enable the detection of LLMs poisoned by our test label and the BigBench canary GUID to, ideally, prevent the inclusion of our labels in the test set in the first place.

## 6 Baselines

In Table 3, we provide baseline results for the test split. Notably, token-based Jaccard similarity performs slightly above random chance, while a simple prompt in GPT-4o-mini shows substantial improvements over the random baseline. In Track B, the *story-emb* system (Hatzel and Biemann, 2024a) only slightly outperforms a simple sentence-similarity encoder model. Interestingly, *story-emb* is outperformed by its foundation model *e5*. We, like Tripathi et al. (2026), attribute this to our task being essentially adversarially designed with respect to *story-emb*: we select triples that are very similar under its understanding of similarity.

| System | Accuracy (%) |
| --- | --- |
| *Track A* | |
| Random | 50.00 |
| Jaccard Similarity | 56.25 |
| GPT-4o-mini | 67.00 |
| *Track B* | |
| Random | 50.00 |
| all-MiniLM-L6-v2 | 58.50 |
| e5 | 66.75 |
| story-emb | 63.25 |

Table 3: Baseline performance results for Track A and Track B

## 7 Participant Results & Leaderboard

Our task attracted 44 participants with paper submissions for Track A alone. Due to space limitations, we list the full results in Appendix A and on our website.[5] The team that initially held the top score on CodaBench, our competition platform, was found to be cheating by using multiple accounts to infer gold labels on the test set and was disqualified. All teams that registered for the final submission phase after the CodaBench competition concluded also submitted papers and are therefore listed here.

We evaluate all submissions in terms of accuracy in a binary classification setup.[6] In Track B, this is done on the basis of embedding cosine distances, while in Track A, the similarity judgment is provided directly.

### 7.1 Track A

For Track A, we show the results for 44 submissions in Table 4 in Appendix A, where the respective systems are cited. A total of 14 teams surpass the 70% accuracy mark. In Track A, the three best-scoring teams are:

- 🥇 In first place, with an accuracy of **78.00%**, *COGNAC* (Erana et al., 2026), an LLM-based system with majority vote across generations of the same model, aspect decomposition for extracting text for the three similarity aspects, and dynamic routing to handle difficult cases (as indicated by within-model, cross-generation vote distribution) separately.

- 🥈 In second place, with an accuracy of **75.75%**, *FactUEP* (Sawinski, 2026), an LLM-based decomposition and scoring system using "weak gating", a form of dynamic routing that prompts the model for reevaluation in cases where aspect-specific similarity deltas with respect to the anchor story between stories A and B are small.

- 🥉 In third place, with an accuracy of **75.00%**, *AI-Monitors* (Tripathi et al., 2026), a majority vote system making use of two prompted LLMs and one embedding model, with the latter, in a Track B-style setup, making decisions on the basis of cosine-distance.

---
[5]https://narrative-similarity-task.github.io/results/

[6]Recall that the task dataset comes with shuffled labels, meaning the positional annotator bias is not observable, and the classes are equally distributed.

The top-scoring system *COGNAC* performs roughly on par with individual annotators under our analysis. In terms of methodology, we observe that many of the best systems, including *COGNAC*, *AI-Monitors*, *YNU-HPCC*, and *JCT*, rely on some variant of ensembling. While *COGNAC* and *YNU-HPCC* employ voting across multiple decisions of the same model (arguably a sampling strategy rather than an ensemble), *AI-Monitors* and *JCT* ensemble across three distinct models.

Based on various difficulty signals, like vote distribution in ensembles, many submissions make use of dynamic routing in multi-stage systems, reserving more elaborate inference procedures for difficult cases. In cases deemed to be difficult, *COGNAC* explicitly models the three similarity aspects, while *CascadeMind* employs a symbolic tie breaker, and *FactUEP* triggers a reevaluation step. Many of the lower-scoring teams use approaches that appear primarily designed for Track B.

While many teams employ variants of chain-of-thought prompting (CoT), only a few teams specifically ablate for it, reporting mixed results. *TeleAI* finds CoT to provide a great improvement over the standard prompt, while *COGNAC* finds it to perform about on par with their alternative component-based scoring at much greater computational cost, *FactUEP* finds extended reasoning to harm their performance, and *TeamCV* finds few-shot prompting to narrowly outperform CoT.

Despite our dataset containing no overtly offensive content, some teams report issues with the safety filters of commercial LLMs (*JCT* and *CITD@UIT*). In private correspondence, a participant specifically mentioned triggering Gemini's content filter for an example in the development split summarizing stories that described same-sex relationships but did not detail sexual acts.[7]

**Symbolic Representations** In terms of symbolic approaches to narrative similarity, only one team relies primarily on symbolic representations for their similarity comparisons: *IITBoys*. They build structured narrative facets from story summaries, creating structured graphs that are, in turn, transformed into embeddings using a Graph Attention Network. *CascadeMind* employs a symbolic approach for hard cases. *COGNAC* reports having also evaluated the narrative functions by Propp (1968) in various setups, but–given their comparatively low scores of up to 63% accuracy–conclude that Propp's functions are a poor fit for representing the plot summaries in our dataset. *Duluth* make use of VerbNet to symbolically model aspects of narrative.

### 7.2 Track B

The results for Track B are shown in Table 5 in Appendix A, where the respective papers for the 27 submissions are cited. In this setup, model performance drops considerably compared to Track A, with the best-performing system in Track B performing about 10 points worse than the best-performing one in Track A. As stated in Section 5.3, the two tracks follow the same distribution, pointing to an inherent increase in difficulty for Track B. This trend is observed across retrieval applications, with cross-encoders (e.g., rerankers) almost universally outperforming bi-encoders (Thakur et al., 2021).

In Track B, the three best-scoring teams are:

- 🥇 In first place, with an accuracy of **72.00%**, *COGNAC* (Erana et al., 2026), relies on embeddings created by the pretrained model `gemini-embedding-001`, operating purely on an extracted course-of-action description.

- 🥈 In second place, with an accuracy of **71.25%**, *YNU-HPCC* (Song et al., 2026), relies on `gemini-embedding-001` with additional All-but-the-Top post-processing (Mu and Viswanath, 2018).

- 🥉 In third place, with an accuracy of **70.50%**, *hits_team* (Zhou et al., 2026), fine-tunes Qwen3-Embedding-8B using both a contrastive loss and additional soft similarity labels.

Many of the Track B systems attempt to encode the explicit components of similarity as defined in Section 3. In terms of embedding size, a plurality of 9 systems use 1,024, with 6 systems using 768. All other systems use embedding sizes between 5,120 and 256. We observe a Pearson correlation of 0.28 between accuracy and embedding size, but note that the best-performing system, *COGNAC*, uses an embedding size of just 3,072 while 6 other teams use larger embedding sizes.

A popular technique, especially in Track B, is what *JCT* coin *Narrative Core Distillation*, the process of producing a new textual representation of the input that focuses on narrative aspects, often the three aspects we introduce. *CophiWue* employ a three-stage process: after first extracting a narrative core of five events they build a global event taxonomy and, in a third step, rephrase the events using this taxonomy. In Table 5 in the Appendix, we refer to this technique as *aspect extraction*.

---

[7]The effect was likely caused by the summary of *Simon vs. the Homo Sapiens Agenda* which expressly describes the protagonist as *16-year-old*.

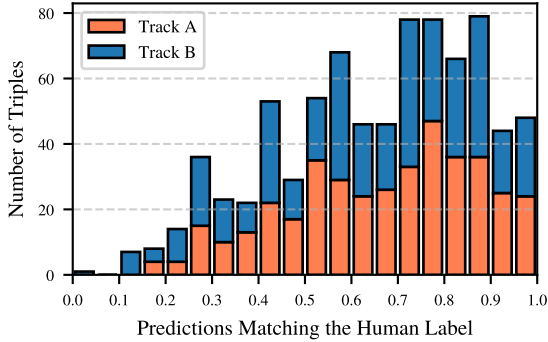

Figure 4: A stacked histogram indicating the distribution of triples based on the fraction of systems that successfully matched the human label, categorized by the track.

We analyze the similarity of individual systems by comparing the Spearman rank coefficient of all their similarity judgments. In Appendix D, we show a hierarchical clustering of all systems informed by the rank coefficient between their respective similarity judgments. Generally, it becomes clear that the best-performing models largely agree in their similarity rankings but we note that they also all make use of either Qwen or Gemini for embeddings. Additionally, we observe that *CuriosAI* and *TFB* are the most similar teams, simply because they both use SentenceT5 embeddings without fine-tuning. In terms of teams that differ in task performance but are similar in ranking, *L3IRIT* and *Narrative Team* stick out: the teams are separated by more than 8 percentage points in terms of task performance, yet share a strong rank correlation of around 0.6. While *Narrative Team* make use of our provided synthetic training data, *L3IRIT* create their own dataset using cross-lingual Wikipedia alignment and both teams rely on different foundation models.

## 8 Data Quality & Annotation Difficulty

To gauge the difficulty of the task, two teams also conducted independent re-annotation studies. *COGNAC* re-annotate the 39 instances from our sample set and qualitatively find many of the samples that are considered difficult under their setup to constitute ambiguous examples. *TFB* annotate 50 examples using three annotators with additional Likert scale confidence estimation. Both studies report inter-annotator agreement comparable to ours ($\alpha = 0.313$ and $\alpha = 0.32$). *COGNAC* also report LLMs achieving a higher agreement with our gold labels than their human re-annotations. *CuriosAI*, in their manual error analysis, find annotations to often be ambiguous and claim that many error cases may be caused by misleading lexical overlap.

Figure 4 illustrates cross-system agreement with the human labels across all instances. Specifically, the stacked histogram visualizes the number of instances–broken down by track–where a given fraction of systems matched the human annotation. For the majority of triples, most systems agree with the human label. Comparing the two tracks, both have a similar number of easy instances where nearly all systems are correct (>0.9). Track B features a noticeably higher number of instances where the vast majority of systems disagree with the label. Generally, triples with very low system agreement may indicate annotation errors, but may also represent instances that are inherently hard to classify. To investigate the potential for annotation mistakes, we manually reinspected challenging cases from Track A. We did not find unambiguous annotation errors, but rather instances that were highly ambiguous (see Appendix E for three examples).

## 9 Quantitative Result Analysis

To gain additional insights, we analyze the submitted decisions and embeddings of all teams.

**Hard Samples** We define hard samples as those that required a third annotation because the initial annotators disagreed. In Track A, across all submitted systems, accuracy on the total of 74 hard samples was just 59.97% while it was 68.05% for the remaining easier samples. The best systems for hard samples are *FactUEP* 71.62% and *AI-Monitors* at 70.27%, both just outperforming the overall best-placed system *COGNAC* at 67.57% by a margin of just two samples. Interestingly, *Mendel292*, ordinarily scoring much lower, also reaches an accuracy of 67.57% on hard samples; we have no convincing explanation for this effect.

**Narrative Embeddings Predict Genre** Guided by the assumption that narratives within the same genre tend to share structural patterns, motifs, and plot dynamics (Frow, 2014; Piata, 2016), we analyze whether narrative representations correspond to genre. Prior work in narratology and computational story modeling suggests that genres con-strain the types of events, character roles, and narrative progressions that occur in stories (Propp, 1968; Chambers and Jurafsky, 2009; Finlayson, 2012). If narrative embeddings capture such structural aspects, we expect stories from the same genre to cluster in the representation space. To this end, we employ a kNN approach (with k=5) and evaluate whether the most frequent genre in a story's k-nearest neighbors is present in its genre list. From this, we derive a genre precision at one (P@1) for each submitted system. We find a strong Pearson correlation of 0.67 between the narrative similarity accuracy and kNN-based genre P@1. These findings suggest that, for the approaches explored in this shared task, narrative is highly associated with genre. In Figure 5, we show the P@1 on a per-system basis. The system by *TeleAI* is a clear outlier; we hypothesize that its overperformance in genre identification is caused by their rewriting strategy for creating synthetic data, but note that they only prompt for shared themes, not genres.

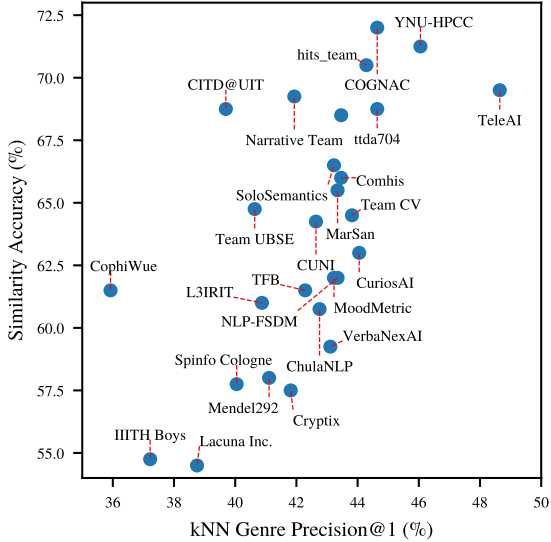

Figure 5: We compare the accuracy in our similarity setup with the kNN-based genre accuracy.

When we employ a uniform kNN regression (i.e., not distance-weighted) to predict the release year of a given story, we observe a small negative Pearson correlation (-0.14) between the system's mean squared error and its narrative similarity accuracy. We can conclude that narrative embeddings in our setup do not predict a story's release year.

**Cross-Team Ensembles** For Track A, majority-vote ensembling across the top 20 systems achieves an accuracy of 79.25%, slightly outperforming even the best individual system. In the case of Track B, we ensemble over the top 5 systems by concatenating L2-normalized vectors and subsequently renormalizing to calculate cosine distances. Under this setup, we achieve an accuracy of 73.75%, again slightly improving on the top-scoring system.

**Positional Biases** While we observe strong positional biases in the human annotators, they are largely not present in the submitted system. In fact, while our (shuffled) gold labels contain 208 instances of text A being closer, the submitted systems only predict A an average of 191 times each out of 400 times. This bias is in large part caused by the team *CICL26*, who predict *A* only 86 times, but still achieve a respectable accuracy of 59%; without this outlier, the average number of A predictions is 196.4. In terms of competitively scoring systems, *JCT* exhibit the greatest deviation from an equal distribution, predicting A just 40.75% of the time. In the past, LLMs were known to exhibit strong positional biases (Shi et al., 2025); the fact that we cannot see any preference for option *A* despite many approaches relying heavily on prompting likely indicates that most modern models no longer exhibit this bias. *CITD@UIT* and *CICL26* report explicitly correcting for positional biases in models by swapping the stories A and B across inference runs.

## 10 Conclusion

With this shared task, we introduce a challenging benchmark that, despite the tasks' inherent subjectivity and corresponding label variation, yields a relatively wide spread in terms of accuracy scores across participants. None of the submitted systems reach our theoretical limit for label quality (see Section 5.3), and cross-participant ensembles outperform the best individual system; both results indicate remaining, albeit limited, room for improvement. While we do not conclusively establish a label quality ceiling, our analysis also shows good final label quality with respect to theoretical oracle labels.

Determining which story is more similar remains subjective under our annotation guidelines, primarily because we do not specify weightings for the individual similarity aspects. We expect the dataset to see future use, particularly for evaluating narrative representation learning. We recommend evaluating on the full set rather than the separate Track A and B splits. Future work in this domain should, in our view, aim to incorporate human label variation. In our released dataset, we provide individual human labels and annotator comments.

In addition to being the first work to provide a large-scale public dataset for narrative similarity in fictional stories and attracting a wide range of participants to our shared task, our approach improves on prior work in two ways. First, our annotations are conducted in a contrastive setup: participants are asked to rate which of two candidate stories is more narratively similar to a given anchor story. We consider this advantageous because scalar judgments are difficult to annotate consistently and lack discriminative value when comparing pairs of roughly equal similarity. Secondly, absent a well-established, all-encompassing definition of narrative similarity, we propose and successfully operationalize an intuition-driven notion that is nevertheless compatible with narrative theory. We push the annotators to provide a largely intuitive judgment rather than one driven by concrete features, which would always be subject to debate in scientific and scholarly communities.

## 11  Limitations

Under certain assumptions, our data approaches a label quality of almost 90% with respect to oracle labels. Crucially, all systems may still be lacking in some respects regarding similarity judgments, and we cannot make a definitive statement about the performance ceiling enabled by our labels." It is to be noted that the idea of oracle labels itself does not fully capture the inherent ambiguity of the task and future research should prioritize human label variation to better quantify this task's inherent ambiguity. Ideally, such an approach would also involve recruiting a diverse set of annotators with respect to socio-cultural and linguistic backgrounds.

Our sampling method based on *story-emb* (Hatzel and Biemann, 2024a) is likely to have introduced a bias in the triples selected for our dataset. Similarly, our rejection filtering means that the dataset may focus heavily on aspects of narrative that today's models do not natively capture.

Our work is purely focused on English, and we hope future work can address this shortcoming. The impact of our language choice likely also extends to the types of stories in our datasets, as storytelling conventions vary vastly across cultures.

## 12  Ethical Considerations

We believe the task of narrative similarity to be a low-risk application for language models, as no content is generated, and we see no immediately harmful applications for narrative similarity judgments.

The compensation provided to annotators was set at at least twice the applicable legal minimum wage in their respective countries of residence; specific pay rates are not disclosed due to Toloka's policies. Annotators are warned about sensitive or potentially upsetting topics they may encounter in the data and are allowed to skip a task if they feel uncomfortable.

## 13  Acknowledgements

The work was supported by the German Research Foundation (DFG) (grant BI 1544/11-2 and GI 1105/3-1) as part of the project "Unitizing Plot to Advance Analysis of Narrative Structure (PLANS)" as part of the priority program "Computational Literary Studies (CLS)" (SPP 2207). We thank all participants in the shared task for their efforts and contributions, and we are grateful to the anonymous reviewers for their valuable feedback and pointing out limitations that we had not yet considered. We thank Toloka AI for providing annotation resources.

## A Full Results

The full results are found in Table 4 for Track A and in Table 5 for Track B.

## B Quality Filter Prompt

We employ the following prompts to select story summaries that meet our quality criteria. The prompts retain some open-ended summaries, but, in our manual evaluation, remove those summaries that do not reveal any of the actual story.

**Developer Prompt**

*You are a quality assurance tool, answer only with 'True' if the quality is sufficient and with 'False' if it is not.*

**User Prompt**

*Is the following text the summary of a story? This means it should not just be a teaser and at least reveal a sizable part of the plot, not just the premise. It should contain minimal Markdown, no headings!*
*Text:*

| Rank | Team | Accuracy | Author | Description |
|---|---|---|---|---|
| 1 | COGNAC | 78.00 | Erana et al. (2026) | LLM majority vote + aspect decomposition + dynamic routing |
| 2 | FactUEP | 75.75 | Sawinski (2026) | LLM + dynamic routing |
| 3 | AI-Monitors | 75.00 | Tripathi et al. (2026) | Three model majority vote with few-shot + embeddings |
| 4 | TeleAI | 74.75 | Fu et al. (2026) | LLM + finetuning with synthetic data |
| 5 | YNU-HPCC | 74.25 | Song et al. (2026) | Embedding/prompting + dynamic routing |
| 6 | UTD-HLTRI | 74.00 | Ailneni et al. (2026) | LLM + chain-of-thought + in-context learning with specifically ordered examples |
| 7 | JCT | 73.75 | Rosenfeld et al. (2026) | Three LLM ensemble |
| 8 | CuriosAI | 73.50 | Shibata et al. (2026) | Zero-shot embedding distance |
| 9 | Yam | 73.00 | Yam and Yam (2026) | LLM with iterative prompt optimization |
| 10 | CascadeMind | 72.75 | Kawada and Holyoak (2026) | LLM majority vote + dynamic routing |
| 11 | Team CV | 70.75 | R S and Ulli (2026) | LLM with few-shot |
| 12 | hermeneutic_hools | 70.50 | Upravitelev et al. (2026) | Cross-persona LLM majority vote |
| 13 | NCL&HKU-NarrSim | 70.25 | Xu et al. (2026) | Aspect Extraction + LLM decisions |
| 13 | NarSiL | 70.25 | Grecu et al. (2026) | LLM majority voting + dynamic routing with aspect analysis |
| 15 | DUTIR | 69.75 | Borjigin and Yang (2026) | Fine-tuned LLM + cleaned training data |
| 16 | CophiWue | 69.50 | Konle and Jannidis (2026) | Zero-shot prompting |
| 17 | ttda704 | 69.25 | Tran and Thien (2026) | Embeddings + synthetic training data |
| 17 | CITD@UIT | 69.25 | Nguyen et al. (2026) | LLM with chain-of-thought |
| 19 | Narrative Nexus | 68.50 | GUO et al. (2026) | Fine-tuned LLM |
| 20 | harapalb | 68.25 | Carp (2026) | LLM + embedding + syntax features |
| 20 | Comhis | 68.25 | Shu et al. (2026) | LLM + embeddings |
| 22 | SoloSemantics | 68.00 | Au (2026) | embeddings |
| 23 | schmerle | 67.75 | Schmerle et al. (2026) | LLM + structured prompting |
| 24 | L3IRIT | 65.75 | Hamdi et al. (2026) | Embedding finetuning with Wikipedia derived data + entity masking |
| 25 | NLP-FSDM | 65.50 | Benlahbib et al. (2026) | Embedding fine-tuning |
| 26 | Team UBSE | 65.25 | Marogel and Popescu (2026) | Aspect extraction + embedding |
| 27 | MoodMetric | 65.00 | Bolisetty et al. (2026) | Fine-tuned embedding model ensemble |
| 27 | MarSan | 65.00 | Najafi et al. (2026) | Fine-tuned embeddings + rewriting |
| 29 | PEU Lab | 64.50 | Ma et al. (2026) | Fine-tuned embeddings + synthetic data |
| 30 | Narrative Team | 64.25 | Khaidukova et al. (2026) | fine-tuned embeddings + synthetic data + ranking objective |
| 31 | blue | 62.50 | Sharma et al. (2026) | Embedding ensemble |
| 32 | ChulaNLP | 63.50 | Gampper et al. (2026) | Aspect extraction (13 aspects with learned weighting) + embedding model |
| 32 | Team HausaNLP | 61.50 | Adam et al. (2026) | TF-IDF |
| 33 | TFB | 61.25 | Colli et al. (2026) | Fine-tuned LLM |
| 34 | Spinfo Cologne | 60.25 | Pagel and Reiter (2026) | Aspect extraction + embeddings |
| 35 | LIAAD INESCTEC | 59.75 | Amorim et al. (2026) | Discourse representation structure (DSR) + embeddings |
| 36 | Cryptix | 59.50 | M et al. (2026) | Embeddings |
| 37 | CICL26 | 59.00 | Zhang and Yu (2026) | LLM majority voting + symmetry test |
| 38 | Duluth | 58.50 | Bevers et al. (2026) | Embedding + dynamic routing + cross-encoder |
| 39 | IIITH Boys | 57.75 | Vijayachandran et al. (2026) | LLM-based story-graph extraction |
| 40 | Lacuna Inc. | 57.25 | Kudelya et al. (2026) | State-space model feature extraction |
| 41 | Mendel292 | 56.50 | Gruppi et al. (2026) | Embedding finetuning + multi-task classification loss |
| 42 | PLlama | 55.50 | Jain (2026) | LLM with CoT |
| 45 | VerbaNexAI | 53.50 | Pertuz-Duran et al. (2026) | Embeddings + classification head |

Table 4: Track A results given as accuracy in percent.

## C Synthetic Story Prompts

We use a variety of models to create more diverse stories, specifically we use: 'gpt-4o-mini', 'gpt-4o', 'gpt-5-chat', 'claude-sonnet-4', 'llama3-1-8b-turbo', 'meta-llama/Meta-Llama-3.1-70B-Instruct-Turbo', 'Qwen/Qwen2.5-7B-Instruct-Turbo', 'Qwen/Qwen2.5-72B-Instruct-Turbo', 'meta-llama/Llama-3.3-70B-Instruct', and 'deepseek-ai/DeepSeek-V3'.

> **Seed Topic Prompt**
>
> Write a short story of 5–8 sentences in the style of a Wikipedia film plot summary. Do not include a title, year, or meta information. Describe only the plot in a neutral, encyclopedic tone, as if summarizing the events of a film. The general topic of the plot is {{seed_word}}.

> **Similar Story Prompt**
>
> Write a new story that closely follows the same context and general storyline. Keep the same sequence of events and plot structure, but change all names of people, places, and specific details of the setting. The result should feel like a different story while clearly mirroring the original plot. Write it in 5–8 sentences, in the style of a Wikipedia film plot summary. Do not include a title, year, or meta information — output the plot only.

> **Distant Story Prompt**
>
> Write a new short story of 5–8 sentences that is only loosely inspired by the original. You may change the setting, characters, genre, tone, and sequence of events. You are encouraged to create a different conflict and resolution, and the ending should not mirror the source. The story should retain only a faint thematic connection or mood from the original, while otherwise standing as a distinctly new narrative. Write it in the style of a Wikipedia film plot summary. Do not include a title, year, or meta information — output the plot only.

## D Track B: Hierarchical Clustering

See Figure 6 for a visualization of the relationship of models' similarity rankings.

## E Challenging Examples for Track A

### E.1 Sample Index 167

Correct Models: 8 | Gold A

>  **Anchor**
>
> Ken and Barbara are tourists on safari in a Borneo jungle, led by tour guide Buck Malone. The group is taken prisoner by a native tribe. While tribal women strip and oil Barbara to prepare her for the chief, Ken is set free so that the chief's men can kill him. The chief shows Barbara Ken's severed head and then rapes her. When Buck is set free, he rescues Barbara and they escape.

> **Text A**
>
> In the mid-1500s, a witch is burned in Scotland and places a curse on the inhabitants before she dies. One hundred years later, the tree she was chained to and burned still stands with no one daring to destroy it. The curse remains by forcing women to commit suicide. The witch's descendant, Martha Gunt, is sentenced to be burned for witchcraft. As she is placed in a prison cell, Maciste comes forth to fight evil. When he uproots the cursed tree, he finds an entrance to Hell where he attempts to track down the original witch to rescind her curse and attempts to help the damned from their plight.

> **Text B**
>
> Beautiful but deadly Lyra the She-Devil and her ivory-hunting friends have discovered a large herd of bull elephants and plot to capture them, forcing an East African native tribe to serve as bearers. Their ivory poaching plans meet opposition when Tarzan gives his deafening jungle cry. The tusked creatures come running, stomping all over Lyra's plans.

### E.2 Sample Index 248

Correct Models: 9 | Gold B

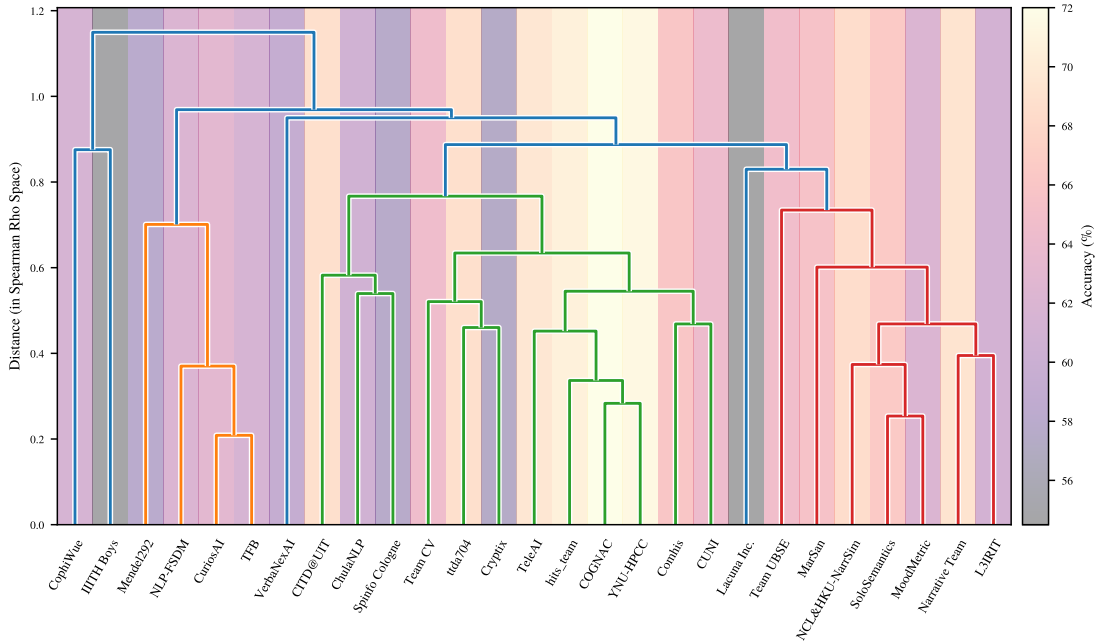

Figure 6: Participant clustered by embedding similarity as measured by pairwise Spearman rank coefficient across all pairs of stories in our Track B test set.

### Anchor

A team of foreign agents arrive in Brazil and discuss an international plot to take control of the country's natural resources. They begin by installing devices on the country's main radio and TV transmitters. The devices are designed to emit subliminal messages geared towards changing the populace into a pliable and aggressive mob supporting a revolution. Journalist Carlos, who is out of the country on assignment when the messages begin, and his friend Laura, who does not listen to the radio or watch television, are unaffected by the broadcasts and both soon realize that something has gone wrong in the country. They begin an investigation and uncover the plot by the agents.

### Text A

The story concerns an experimental nuclear cruise missile which is stolen from a Soviet military site in the USSR. An international terrorist group, under the command of a European power-crazed man known only as the Baron is responsible. The Baron plots to use the stolen Soviet missile to destroy an international peace conference in one week located on an island in the Persian Gulf. When the U.S. consul to Iran is murdered by the Baron's henchmen, Alec Franklin, a US intelligence agent, is ordered to travel to Iran to take over as consul as well as investigate the murder. Upon arrival in Tehran, Alec is followed by two of the Baron's henchmen who attempt to kill him, but Alec manages to escape. Alec then travels from Tehran to Abadan where he meets Konstanine, a Soviet KGB intelligence agent who is in Iran searching for leads to locate the missing cruise missile, which leads to Alec and Konstantine joining forces along with Galina, another Soviet agent, and Leila, an undercover Iranian policewoman, to investigate the Baron in order to find the location to where the cruise missile is being kept before it is used to start World War III.

### Text B

Supercriminal Dr. Fu Manchu plots to freeze the world's oceans with a diabolical new device. With his beautiful but evil daughter, Lin Tang, his army of dacoits, and the help of the local crime organization led by Omar Pasha (whom Dr. Fu Manchu double-crosses), Dr. Fu Manchu takes over the governor's castle in Istanbul, which has a massive opium reserve, to control the largest opium port in Anatolia, since the drug is an important ingredient for the fuel for his machine. Dr. Fu Manchu needs the help of an intelligent scientist with an ailing heart whom he has imprisoned. In order to keep the scientist alive, he kidnaps a doctor and his wife to give the scientist a heart transplant from one of his obedient servants. Opposing him from Britain's branch of Interpol are his nemeses, Nayland Smith and Dr. Petrie.

## E.3 Sample index 251

Correct Models: 8 | Gold B

###  Anchor

The prophet Al Mustafa has lived in the city of Orphalese for 12 years and is about to board a ship which will carry him home. He is stopped by a group of people, with whom he discusses topics such as life and the human condition. The book is divided into chapters dealing with love, marriage, children, giving, eating and drinking, work, joy and sorrow, houses, clothes, buying and selling, crime and punishment, laws, freedom, reason and passion, pain, self-knowledge, teaching, friendship, talking, time, good and evil, prayer, pleasure, beauty, religion, and death.

### Text A

A young workman from Tuttlingen (then part of the Duchy of Württemberg) visited the cosmopolitan city of Amsterdam for the first time in his life and was impressed by a particularly stately home and a large ship laden with precious commodities. He innocently asked people about the owners of the house and the boat and both times the answer was "Kannitverstan", which means "I can not understand you". The simple-minded workman, however, believed that it was the name of a man called "Kannitverstan", and was impressed by the supposed Mr. Kannitverstan's wealth, and at the same time felt victimized in the face of his own poverty. Later in the day, he observed a funeral procession and asked one of the mourners who the deceased was. When he received the answer "Kannitverstan" he mourned for the late Mr. Kannitverstan, but at the same time felt very light-hearted, because he realized that death knows no social differences and everything in life is fleeting. Thus, the workman suffered his own poverty much better.

### Text B

Bruno (Celhay) is a 35-year-old architect who leads a seemingly perfect life. He resides in a beautiful house with his wife and child and also owns a thriving architecture office. However, despite his comfortable lifestyle, he experiences a profound sense of unease. As a result, he decides to leave everything behind and move out to live alone, coincidentally when a businessman approaches him to design an icon for the city of Santiago. With newfound motivation, he embarks on a quest to search for heritage traces, accompanied by Fer (Emilio Edwards), a 29-year-old, energetic, and captivating gay history professor.

| Rank | Team | Accuracy | Author | Description |
| --- | --- | --- | --- | --- |
| 1 | COGNAC | 72.00 | Erana et al. (2026) | Gemini embedding model + Aspect extraction (just course-of-action) |
| 2 | YNU-HPCC | 71.25 | Song et al. (2026) | Gemini embedding model + All-but-the-Top |
| 3 | hits_team | 70.50 | Zhou et al. (2026) | Fine-tuned Qwen3 embedding model + soft similarity labels |
| 4 | TeleAI | 69.50 | Fu et al. (2026) | Fine-tuned Qwen2 + synthetic data generation |
| 5 | Narrative Team | 69.25 | Khaidukova et al. (2026) | Fine-tuned RoBERTa + synthetic data |
| 6 | ttda704 | 68.75 | Tran and Thien (2026) | Fine-tuned MPNet-base + aspect extraction + aspect views |
| 6 | CITD@UIT | 68.75 | Nguyen et al. (2026) | Fine-tuned BGE-M3 + failure informed synthetic data |
| 8 | NCL&HKU-NarrSim | 68.50 | Xu et al. (2026) | Fine-tuned RoBERTa + self-supervised similarity training |
| 9 | SoloSemantics | 66.50 | Au (2026) | Fine-tuned MPNet-base + chunking |
| 10 | Comhis | 66.00 | Shu et al. (2026) | Qwen3 + fine-tuned projection head + hard example mining |
| 11 | MarSan | 65.50 | Najafi et al. (2026) | Fine-tuned BGE-large + soft labels + data augmentation |
| 12 | Team UBSE | 64.75 | Marogel and Popescu (2026) | Fine-tuned QZhou-Embedding + Aspect extraction + weighting and embedding post-processing |
| 13 | Team CV | 64.50 | R S and Ulli (2026) | OpenAI text-embedding-3 |
| 14 | CUNI | 64.25 | Mitka and Helcl (2026) | Fine-tuned Qwen3 + aspect views + synthetic data |
| 15 | CuriosAI | 63.00 | Shibata et al. (2026) | SentenceT5 no fine-tuning |
| 16 | NLP-FSDM | 62.00 | Benlahbib et al. (2026) | BGE-large (no fine-tuning, see Track A for fine-tuned version) |
| 16 | MoodMetric | 62.00 | Bolisetty et al. (2026) | Fine-tuned BGE-large |
| 18 | TFB | 61.50 | Colli et al. (2026) | SentenceT5 no fine-tuning |
| 18 | CophiWue | 61.50 | Konle and Jannidis (2026) | Aspect extraction + taxonomy inference + controlled vocabulary generation |
| 20 | L3IRIT | 61.00 | Hamdi et al. (2026) | Fine-tuned MPNet + Wikipedia cross-lingual training data + entity masking |
| 21 | ChulaNLP | 60.75 | Gampper and Rutherford (2026) | Aspect extraction (13 aspects with learned weighting) + embedding model |
| 21 | VerbaNexAI | 59.25 | Pertuz-Duran et al. (2026) | Qwen3-0.6B + learned projection head |
| 22 | Mendel292 | 58.00 | Gruppi et al. (2026) | Fine-tuned MiniLM + synthetic data |
| 23 | Spinfo Cologne | 57.75 | Pagel and Reiter (2026) | Aspect Extraction + MPNet |
| 24 | Cryptix | 57.50 | M et al. (2026) | MPNet |
| 25 | IIITH Boys | 54.75 | Vijayachandran et al. (2026) | MPNet + Graph Attention Network |
| 26 | Lacuna Inc. | 55.50 | Kudelya et al. (2026) | Jamba-1.5 + embedding extraction |

Table 5: Track B results given as accuracy in percent.